\documentclass{ifacconf}

\makeatletter
\let\old@ssect\@ssect %
\makeatother

\usepackage{natbib}
\usepackage{hyperref}
\makeatletter
\def\@ssect#1#2#3#4#5#6{%
	\NR@gettitle{#6}%
	\old@ssect{#1}{#2}{#3}{#4}{#5}{#6}%
}
\makeatother
\usepackage{graphicx}

\usepackage{booktabs} %
\usepackage{xcolor}

\usepackage{amsmath}
\usepackage{amssymb}
\usepackage{mathtools}

\newcommand{\quat}[0]{\ensuremath{\mathbf{q}}}

\newcommand{\quatFromTo}[2]{\ensuremath{\prescript{#1}{#2}{\quat}}}

\newcommand{\sca}[1]{\ensuremath{#1}}

\newcommand{\scadefN}[1]{\ensuremath{\sca{#1} \in \mathbb{N}}}
\newcommand{\ve}[1]{\ensuremath{\boldsymbol{#1}}}

\newcommand{\vecdefN}[2]{\ensuremath{\ve{#1} \in \mathbb{N}^{#2}}}

\newcommand{\mat}[1]{\ensuremath{\mathbf{#1}}}
\newcommand{\matOnlydefR}[2]{\ensuremath{\in \mathbb{R}^{#1 \times #2}}}
\newcommand{\matdefR}[3]{\ensuremath{\mat{#1} \matOnlydefR{#2}{#3}}}
\newcommand{\matdefRThreeD}[4]{\ensuremath{\mat{#1} \in \mathbb{R}^{#2 \times #3 \times #4}}}

\newcommand{\matdefH}[3]{\ensuremath{\mat{#1} \in \mathbb{H}^{#2 \times #3}}}

\newcommand{\code}[1]{\ensuremath{\mathtt{#1}}}

\DeclareMathSymbol{\dd}{\mathord}{operators}{"3A}

\newcommand{\ind}[1]{\ensuremath{[#1]}}

\newcommand{\FromToIndFromTo}[5]{\ensuremath{[#1 \dd #2, #3, #4 \dd #5]}}

\newcommand{\fromToInd}[3]{\ensuremath{[#1 \dd #2, #3]}}

\usepackage[separate-uncertainty=true]{siunitx}

\usepackage{threeparttable}

\usepackage{pifont}%
\definecolor{darkgreen}{RGB}{51, 153, 51}
\definecolor{lightred}{RGB}{255, 102, 102}
\newcommand{\cmark}{\textcolor{darkgreen}{\ding{51}}}%
\newcommand{\xmark}{\textcolor{lightred}{\ding{55}}}%

\usepackage{url}

\begin{document}
\begin{frontmatter}

\title{Dispelling Four Challenges in Inertial Motion Tracking with One Recurrent Inertial Graph-based Estimator (RING)} 

\author[First]{S. Bachhuber} 
\author[First]{I. Weygers} 
\author[Second]{T. Seel}

\address[First]{Department Artificial Intelligence in Biomedical Engineering, FAU Erlangen-Nürnberg, 91052 Erlangen, Germany (e-mail: first.last@fau.de)}
\address[Second]{Institute of Mechatronic Systems, Leibniz Universität Hannover, 30167 Hannover, Germany (e-mail: first.last@imes.uni-hannover.de)}

\begin{abstract}                %
In this paper, we extend the Recurrent Inertial Graph-based Estimator (RING), a novel neural-network-based solution for Inertial Motion Tracking (IMT), to generalize across a large range of sampling rates, and we demonstrate that it can overcome four real-world challenges: inhomogeneous magnetic fields, sensor-to-segment misalignment, sparse sensor setups, and nonrigid sensor attachment. RING can estimate the rotational state of a three-segment kinematic chain with double hinge joints from inertial data, and achieves an experimental mean-absolute-(tracking)-error of $8.10\pm 1.19$ degrees if all four challenges are present simultaneously.
The network is trained on simulated data yet evaluated on experimental data, highlighting its remarkable ability to zero-shot generalize from simulation to experiment.
We conduct an ablation study to analyze the impact of each of the four challenges on RING's performance, we showcase its robustness to varying sampling rates, and we demonstrate that RING is capable of real-time operation.
This research not only advances IMT technology by making it more accessible and versatile but also enhances its potential for new application domains including non-expert use of sparse IMT with nonrigid sensor attachments in unconstrained environments.
\end{abstract}


\end{frontmatter}

\section{Introduction}
Numerous recent developments in biomedical engineering applications require precise estimation of the motion of articulated bodies in space. Some prominent examples include unobtrusive human motion tracking outside the lab \citep{villa23}, and realizing intelligent symbiosis between humans and robots that enter immersive environments \citep{dafarraEmersive24}. Inertial Measurement Units (IMUs) are used in all these systems because of their unique ability to track movements of articulating rigid bodies of Kinematic Chains (KCs), in a cheaper and more reliable way than State-Of-The-Art (SOTA) multi-camera systems that require continuous line of sight.

All IMU-based motion tracking use cases heavily rely on Inertial Motion Tracking (IMT) algorithms that fuse different measurement signals to estimate motion parameters. This, however, is inherently limited by the following four key challenges \citep{GarVilla2023}: \\
(1) Inhomogeneous magnetic fields indoors and near ferromagnetic materials or electric devices;\\
(2) Sensor-to-segment alignment that involves determining joint positions and axis directions in local sensor coordinates;\\
(3) Solving sparse problems where some segments of the KC are not equipped with a sensor;\\
(4) Addressing real-world disturbances due to nonrigid sensor attachment and caused by large acceleration signals from impacts and soft tissue artifacts.\\
\begin{figure}
    \centering
    \includegraphics[width=.40\textwidth]{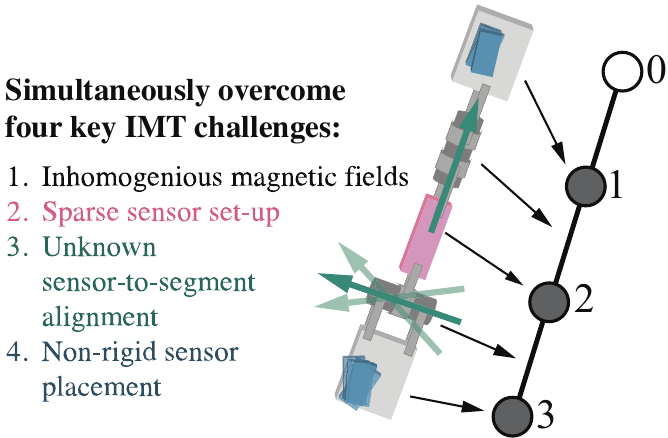}
    \caption{A three-segment KC with two IMUs (blue boxes). The graph representation of the KC is given by the parent array $\ve{\lambda} = (0, 1, 2)^\intercal$. The neural network-based multiple-IMU sensor fusion algorithm RING receives the graph representation and IMU data as input and estimates the rotational state of the KC, overcoming all four key challenges of IMT simultaneously.}
    \label{fig:sys}
\end{figure}
In recent years, many highly specialized methods have been proposed to address these challenges. We will provide a brief overview of the latest and most notable developments accompanied by recent comprehensive methodological overviews. First, a multitude of different kinematic constraints are proposed to replace missing magnetometer heading information, as reviewed in \cite{weygers2023we}. Furthermore, recent general-purpose~\citep{caruso21} magnetometer-free attitude estimators by \cite{laidigVQFHighlyAccurate2023} achieved remarkable accuracy improvements in comparison with, e.g., the widely used filters from \cite{madgwickEfficientOrientationFilter} and \cite{mahonyNonlinearComplementaryFilters2008}. Second, several algorithms have been developed to achieve automatic sensor-to-segment alignment, as outlined by \cite{VitaliPerkin2020} for specific joints with full sensor setups \citep{taetzSelfCalibratingInertialBody2016,mcgrathAutoCalibratingKneeFlexionExtension2018,weygersInvitro}.
Third, a recent trend in sparse sensor setups is visible with methods that either use a limited number of sensors but include magnetometer measurements \citep{syEstimatingLowerLimb2020,syEstimatingLowerLimb2021,huangDeepInertialPoser2018,vonmarcardSparseInertialPoser2017,zhengTrainingDataSelection2021} or are magnetometer-free \citep{grapentinSparseMagnetometerFreeRealTime2020,bachhuberPlugandPlaySparseInertial2023,vanwouweDiffusionInertialPoser2023,yiTransPoseRealtime3D2021,yiPhysicalInertialPoser2022}. Finally, the literature on IMT methods to overcome real-world disturbances is limited and focuses on late interception by outlier rejection techniques (\cite{RemmerKok21}) or further advances in connection constraints \citep{villa21}.

Real-world IMT applications typically present multifaceted challenges, requiring data-driven state observers like the Recurrent Neural Network-based Observer (RNNO) (\cite{bachhuberPlugandPlaySparseInertial2023}) that can effectively address the increasing complexity. To overcome a redundant implementation task in retraining RNNOs for every combination in a large grid of IMT challenges, we proposed the Recurrent Inertial Graph-based Estimator (RING) (\cite{bachuberRINGicml24}) as a pluripotent approach that solves IMT Problems (IMTPs) of tree-structured systems.

Despite RING's ability to provide a solution to a variety IMT challenges, its real-world applicability for a combination of all the aforementioned IMT challenges has not been investigated and their individual impact are unknown. Furthermore, while RING is aimed to be applicable in a plug-and-play fashion, it still requires a specific fixed sampling rate, which vastly limits its applicability in practice. Moreover, the real-time capability in inference has not been explored. In this work, we enhance and validate RING's usability with the following contributions:

\begin{enumerate}
    \item Extending RING's usability by enabling applicability to data from a broad range of sampling rates.
	
    \item Solve \textit{for the first time} the four IMT challenges at once.
	
    \item Show zero-shot experimental generalizability in an extensive ablation study to gain insights on the performance of RING on individual IMT challenges. 
	
    \item Analyze the real-time capability of RING.

\end{enumerate}

\section{Problem Formulation}\label{ch:problem_formulation}
Consider a KC with three segments that are connected by hinge joints with arbitrary and unknown joint axes directions.
Only the outer bodies are equipped with nonrigidly attached IMUs.
A KC is a rigid-body system and it consists of multiple rigid objects (segments) that are rigidly attached to coordinate systems (bodies).
In general, the topology of such a rigid-body system can be represented by a Connectivity Graph (CG) \citep{featherstoneRigidBodyDynamics2008} where nodes represent bodies and edges represent degrees of freedom in the system.
Here, for each segment there is one body with one segment attached to it, such that there is a one-to-one correspondence between segments and bodies.
After the $N$ bodies have been numbered, the CG can be encoded via a parent array $\vecdefN{\lambda}{N}$ where \ve{\lambda}\ind{i} is the body number of the parent of body \sca{i}.
For a three-segment KC, this graph representation and the one parent array utilized in this work is shown in Figure~\ref{fig:sys}.

In this work, the goal is to estimate the complete rotational state of the KC up to a global heading offset \citep{weygers2023we}. We approach this through a filtering problem formulation,  where an estimate of the complete rotational state $\ve{x}(t)$ is obtained at every time instant $t$ from the current and all previous IMU measurements $\ve{y}(t' \leq t)$ that are combined into one measurement signal $\ve{y}(t) \in \mathbb{R}^{12}$ defined as
\begin{equation}
    \boldsymbol{y}(t) = \Big(
    \ve{\omega}{}_1(t)^\intercal, 
    \ve{\rho}{}_1(t)^\intercal, 
    \ve{\omega}{}_3(t)^\intercal, \ve{\rho}{}_3(t)^\intercal \Big)^\intercal \quad \forall t 
    \label{eq:y_t}
\end{equation}
where $\ve{\omega}{}_i(t), \ve{\rho}{}_i(t)$ denote gyroscope and accelerometer measurements at time $t$ of the IMU that is nonrigidly attached to body $\mathcal{S}_i \in \{1, 3\}$.
The rotational state $\ve{x}(t) \in \mathbb{H}^{3}$ of the KC is straightforwardly defined by
\begin{equation}
    \ve{x}(t) = \Big(\quatFromTo{1}{0}(t)^\intercal, \quatFromTo{2}{1}(t)^\intercal, \quatFromTo{3}{2}(t)^\intercal\Big)^\intercal \quad \forall t
 \label{eq:x_t}
\end{equation}
where $\quatFromTo{i}{j}(t)$ denotes the orientation from body $\mathcal{S}_i$ to body $\mathcal{S}_j$ at time $t$ and where body $0$ denotes the earth frame.
Note that the \quatFromTo{1}{0} can only be estimated up to a heading offset from 6D measurements.

Real-world applicability requires solving all of the following challenges of the IMTP that is said to
\begin{itemize}
    \item be \textbf{magnetometer-free} (or 6D in contrast to 9D) if the IMUs measure only three-dimensional angular rates and specific forces, and not provide magnetometer readings.
    \item require \textbf{sensor-to-segment alignment} when hinge joint axes directions are unknown.
    \item be \textbf{sparse} if not every segment that constitute the KC has an IMU attached. Here,the middle segment does not have an IMU attached. An IMTP that has an IMU attached to each body is said to have a full IMU setup.
    \item suffer from \textbf{motion artifacts} if the IMUs are not rigidly attached to the respective bodies, such that there can occur transnational and rotational motions between the segment and IMU. An IMTP without motion artifacts assumes that there cannot exist any relative motion between segment and IMU.
\end{itemize}
From this, it follows that the IMTP considered here is magnetometer-free, requires sensor-to-segment alignment, sparse, and suffers from motion artifacts.

\section{Methods}\label{ch:methods}
In this work, we extend the Neural Network-based (NN-based) multiple-IMU sensor fusion algorithm from \cite{bachuberRINGicml24}.
We address all four key IMT challenges outlined in Section~\ref{ch:problem_formulation}, while enabling sampling rate robustness and showcase that RING is real-time capable. This is achieved by both adapting the training procedure (Section: \ref{ch:rcmg}) and the NN-based multiple-IMU sensor fusion algorithm (Section: \ref{ch:ring}).

\subsection{Simulated Training Data at Various Sampling Rates}\label{ch:rcmg}
RING~\citep{bachuberRINGicml24} is trained on large amounts of simulated input-output data at various sampling rates. The procedure that generates the data for the training of RING is called the Random Chain Motion Generator (RCMG) \cite{bachuberRINGicml24}.
It generates extensively augmented random motions of KCs with:
\begin{enumerate}
    \item different number of segments,
    \item randomized segment lengths,
    \item randomized IMU placement,
    \item randomized joint axes directions, and
    \item rigidly or nonrigidly attached IMUs (by simulating spring-damper-systems with randomized damping and stiffness parameters)~\citep{bachuberRINGicml24}.
\end{enumerate}
From these random KC motions we compute IMU and orientation measurements, but in this work at various sampling rates. These form the input-output pairs for training RING.

RCMG can be summarized as a function that only from PseudoRNG returns the training pair
\begin{itemize}
    \item \matdefRThreeD{X}{T}{N}{10}, where $\mat{X}\FromToIndFromTo{}{}{i}{}{6}$ is the 6D IMU data for body $i$ (if it is not dropped out), and where $\mat{X}\FromToIndFromTo{}{}{i}{6}{9}$ is the joint axis direction of the hinge joint between body $i$ and its parent (if it is not dropped out, and if the parent is not the base), and where $\mat{X}[\dd, i, 10]$ is the inverse sampling rate $\frac{1}{F}$, and
    \item \matdefH{Y}{T}{N} where $\mat{Y}\fromToInd{}{}{i}$ is the orientation from body $i$ to its parent $\ve{\lambda}[i]$, and
\end{itemize}
where $T$ is the number of timesteps, and $N$ is the number of bodies in the KC (here $N=3$).

To achieve a wide coverage, training data is generated for sampling rates drawn from 
$$F \in \{40, 60, 80, 100, 120, 140, 160, 180, 200\} \qty{}{\hertz}$$ 
and to allow for efficient training data batching, the sequence duration is adjusted based on the sampling rate to achieve a common number of timesteps of $T=6000$. A training batch is then built up by stacking $512$ sequences, and additional details regarding the RCMG can be found in \cite{bachuberRINGicml24}.

\subsection{Neural Network Architecture: RING with Sampling Rate Input}\label{ch:ring}
We use a NN trained on data generated using the procedure described in Section~\ref{ch:rcmg}.
The network architecture is based on RING~\citep{bachuberRINGicml24}, a powerful multiple-IMU sensor fusion algorithm that is composed of a decentralized network of message-passing Recurrent NNs (RNNs). Most notably, RING's parameters are defined on a per-node level and shared across all nodes in the graph.
This design enables RING to be applied to broad range of IMTPs with a single set of parameters and enables its exceptional pluripotency.

In this work, the architecture of RING is extended to additionally accept a sampling rate input, such that the dimensionality of RING's per-timestep and per-node input increases by one.
To summarize, RING can be viewed as the following step function that maps the previous state of RING $\ve{\xi}_{t\text{-}1} \matOnlydefR{N}{2H}$ and network input $\mat{X}_t \matOnlydefR{N}{10}$ at time $t$ to the next RING state $\ve{\xi}_t$ and output $\mat{\hat{Y}}_t \in \mathbb{H}^N$, i.e.,
\begin{equation}
\ve{\xi}_t, \mat{\hat{Y}}_t = \code{ring}\left(\ve{\xi}_{t\text{-}1}, \mat{X}_t, \ve{\lambda}\right)
\quad \forall t
\label{eq:ring}
\end{equation}
where \scadefN{H} is the hidden state dimensionality, $\ve{\xi}_0 = \ve{0}$, and where the vector \ve{\lambda} is the parent array, defined in Section~\ref{ch:problem_formulation}.
Internally, RING has the parameters of
\begin{itemize}
    \item the Message-MLP-network $f_\theta: \mathbb{R}^H \rightarrow \mathbb{R}^M$, and
    \item the Stacked-GRUCell-network $g_\theta: \mathbb{R}^{2H} \times \mathbb{R}^{2M+10} \rightarrow \mathbb{R}^{2H}$ which consists of the sequence of Gated-Recurrent-Unit(GRU)Cell, LayerNorm, GRUCell \citep{choLearningPhraseRepresentations2014}, and
    \item the Quaternion-MLP that combines a Layernorm, and a MLP-network $h_\theta: \mathbb{R}^H \rightarrow \mathbb{R}^4$, and
\end{itemize}
where \scadefN{M} is the dimensionality of the messages that are passed along the edges of the graph.
Note that the two GRUCells each have a hidden state dimensionality of $H$, thus the hidden state of RING is of dimensionality of $2H$.
Then, equation \eqref{eq:ring} consists of several consecutive steps, for all $N$ bodies:
\begin{enumerate}
    \item Messages $\mat{M}_t \matOnlydefR{N}{M}$ are computed using $f_\theta$.
    \item Messages are passed along the edges of the graph.
    \item The hidden state is updated using $g_\theta$.
    \item The unnormalized output \matdefR{\tilde{Y}}{N}{4} is computed using the Quaternion-MLP.
    \item The output is normalized to allow interpretation as unit quaternions. The final RING output is one unit quaternion per node $\mat{\hat{Y}}_t \in \mathbb{H}^N$.
\end{enumerate}

RING is trained by comparing $\mat{\hat{Y}}_t$ to the ground truth $\mat{Y}_t$ which is provided by the RCMG, and by minimizing the mean-squared orientation error.

Additional details regarding the the RING architecture and its optimization strategy can be found and is exactly the same as in \cite{bachuberRINGicml24}.

A software implementation of the RCMG and RING, and the source code for creating the results presented in Section~\ref{ch:exp_val} are hosted on GitHub\footnote{\url{https://github.com/SimiPixel/ring}}.

\section{Results and Discussion}\label{ch:results}
In this section, we evaluate the accuracy of RING's orientation estimation, trained in simulation only, when applied to experimental data (see Section~\ref{ch:exp_data}) of the problem specified in Section~\ref{ch:problem_formulation}.
The performance of RING is compared to (SOTA) methods (see Section~\ref{ch:exp_val}).

\emph{
It is remarkable, that RING can solve an experimental IMTP that combines all four challenges in IMT (magnetometer-free, unknown sensor-to-segment alignment, sparse sensor setup, and nonrigid sensor attachment) simultaneously, despite being trained on simulated data only.
}

\begin{figure}
    \centering
    \includegraphics[width=.48\textwidth]{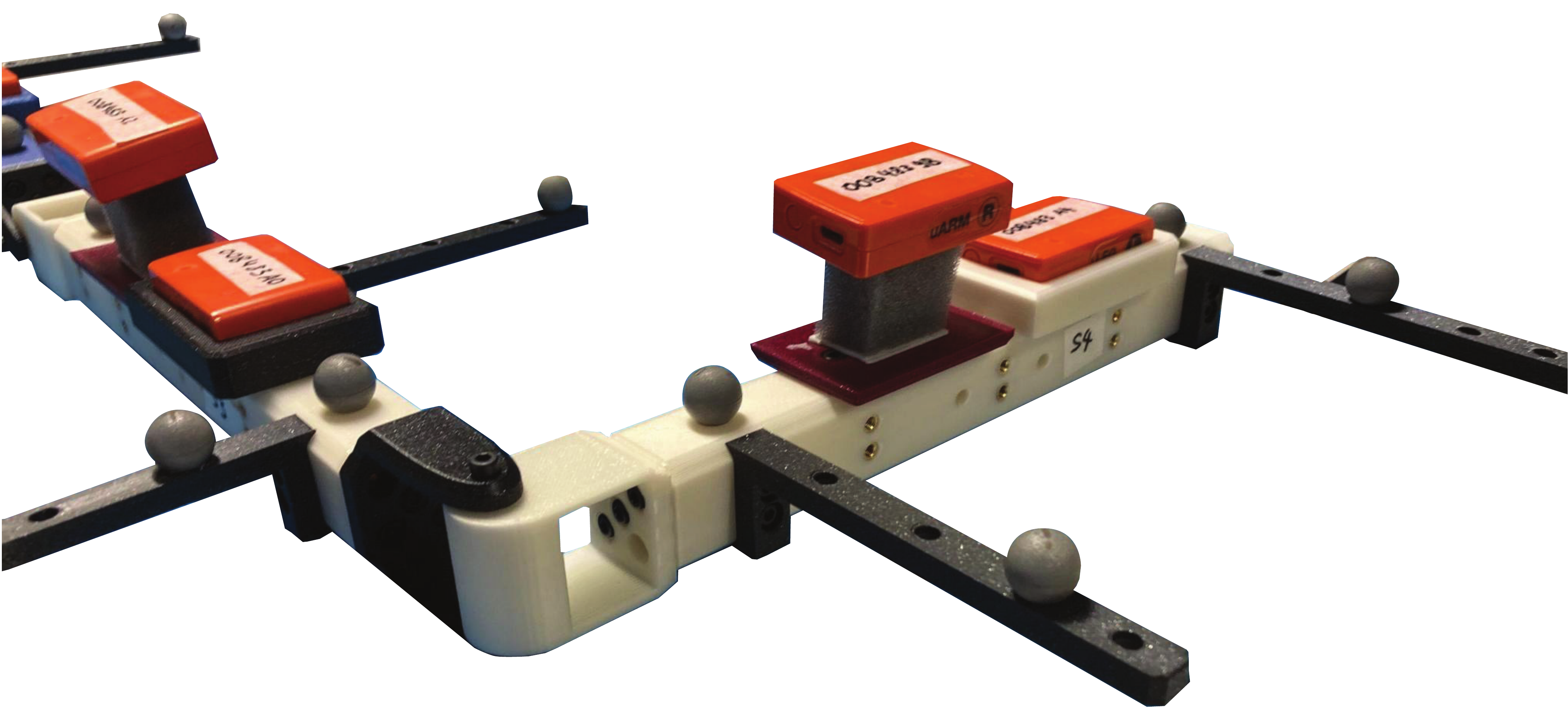}
    \caption{Experimental 3D-printed KC used to validate the RING algorithm. To validate that RING overcomes the IMT challenge of non-rigid sensor placement, each segment of the KC has an IMU attached using foam padding. Additionally, a second IMU is rigidly attached to assess the impact of the foam padding on the accuracy of orientation estimates. Figure from \cite{bachuberRINGicml24}.}
    \label{fig:foam}
\end{figure}

\subsection{Experimental Setup and Data Acquisition}\label{ch:exp_data}
We utilize a five-segment KC to record the experimental data but only the data for the given IMTP relevant parts of the KC are used for evaluation. The five-segment KC includes a singular spherical joint followed by three hinge joints, each oriented along the x, y, and z axes, respectively. Each segment of the KC was equipped with two IMUs: one rigidly attached to the segment and another nonrigidly attached using foam padding, as depicted in Figure~\ref{fig:foam}.

Two distinct trials were conducted, involving random movements of the five-segment KC. Here, two three-segment KCs are thus extracted, one with joint axes direction x and y, and one with y and z.
During evaluation, the first trial spans a duration of \qty{66}{\second} and features a diverse range of motions.
Additionally, the second trial, with a length of \qty{68}{\second}, includes periods where the entire KC remained stationary.
Overall, this results in four trials in total.

We refer to~\citep{bachuberRINGicml24} for additional details regarding the experimental setup and preprocessing.

\begin{table}[hb]
\begin{center}
\caption{Experimental magnetometer-free orientation estimation accuracy (in degrees) of RING compared to two SOTA methods. The IMUs are nonrigidly attached and to counteract this influence both VQF and RNNO are used in combination with a low-pass filter with optimized cutoff-frequency. All methods are evaluated at \qty{100}{\hertz}.}
\label{tab:main_results}
\begin{threeparttable}
\begin{tabular}{lcccc}
Method & S. Misal.\textsuperscript{1} & 6D & Sparse & MAE [deg] \\ \toprule
VQF\textsuperscript{2} & \cmark & \xmark & \xmark & $18.45\pm 9.10$\\
RNNO\textsuperscript{3} & \xmark & \cmark & \cmark & $\phantom{1}8.64\pm 4.13$\\
RING & \cmark & \cmark & \cmark & $\phantom{1}8.10\pm 1.19$\\ \bottomrule
\end{tabular}
\begin{tablenotes}
\scriptsize \normalfont
\item \textsuperscript{1} Sensor-to-segment Misalignment
\item \textsuperscript{2} \cite{laidigVQFHighlyAccurate2023}
\item \textsuperscript{3} \cite{bachhuberPlugandPlaySparseInertial2023}
\end{tablenotes}
\end{threeparttable}
\end{center}
\end{table}

\begin{figure*}
    \centering
    \includegraphics[width=\textwidth]{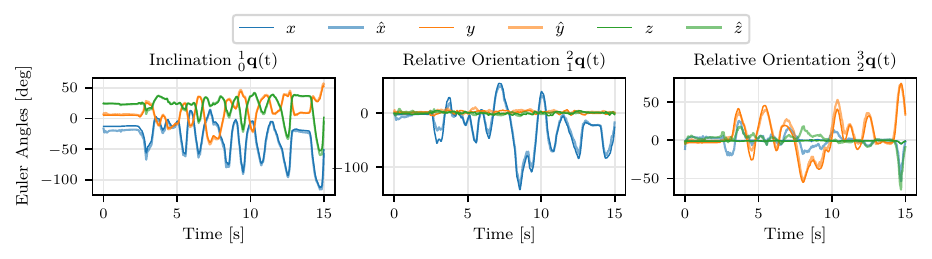}
    \caption{Experimental example sequence that demonstrates RING's prediction performance from sparse, magnetometer-free, nonrigidly attached IMUs and without joint axes direction, and comparing to ground truth orientations for the first \qty{15}{\second} of one trial and for a double hinge joint KC with joint axes directions in $x-$ and $y-$direction.}
    \label{fig:exemplary}
\end{figure*}

\subsection{Evaluation Metrics and Baselines}\label{ch:exp_refs}
The ground truth orientations for the experimental trials (see Section~\ref{ch:exp_data}) were recorded using optical motion capture.
Orientation estimation accuracy is quantified using the Mean-Absolute-(tracking)-Error (MAE) in degrees.
Here, the mean calculation reduces the dimensions of the different trials, time, and three orientations (including inclination and two relative orientations).
In the time dimension, initial \qty{5}{\second} of each trial were deliberately excluded from the MAE calculations. This decision was made to ensure that the recorded errors accurately reflected the system's performance post-convergence.

To the best of the authors' knowledge, there exists no alternative method that can be applied to the IMTP as described in Section~\ref{ch:problem_formulation}.
However, two SOTA baseline methods can be identified after simplifying the IMTP so that it does not contain all four challenges simultaneously.
The first baseline is obtained by using conventional IMT methods, that is, using a full 9D IMU setup and tracking each segment independently.
The SOTA method for such single-IMU sensor fusion is VQF \cite{laidigVQFHighlyAccurate2023}.
The second baseline is obtained after eliminating the challenge of anatomical calibration.
Under the assumption of known joint axes direction, RNNO can be applied \cite{bachhuberPlugandPlaySparseInertial2023}.
Note that since two KCs with different directions of the joint axes are used for experimental validation (see Section~\ref{ch:exp_data}), two trained RNNO networks are required.
To compensate for the violation of the rigid-IMU-attachment assumption, both baselines additionally utilize a low-pass filter.
The cutoff frequency was grid searched and we report only the best result for each baseline method.

\subsection{Experimental Validation of RING}\label{ch:exp_val}
The trained RING is applied to experimental data from an IMTP that combines the four challenges of nonrigid IMU attachment, misaligned sensors and segments, magnetometer-free measurements, and a sparse sensor setup.
The MAE in the orientation estimate for RING and the two SOTA baseline methods are reported in Table~\ref{tab:main_results} and confirms that RING outperforms both alternative methods despite solving the more challenging IMTP.
The first 15 seconds of one example sequence are shown in Figure~\ref{fig:exemplary} and demonstrate RING's prediction performance and quick convergence.

In Table~\ref{tab:ablation}, we conduct an ablation study to analyze the impact of nonrigid IMU attachment, sensor-to-segment alignment, and sparse IMU setup on RING's orientation estimation accuracy.

\begin{figure}
    \centering
    \includegraphics[width=0.48\textwidth]{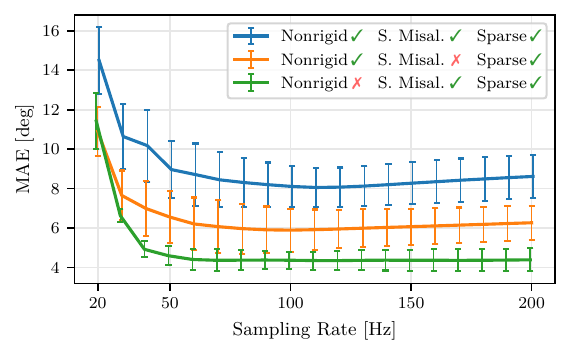}
    \caption{Experimental magnetometer-free motion tracking accuracy (in degrees) of RING across various sampling rates. RING achieves a nearly constant estimation accuracy across sampling rates ranging from \num{50} to \qty{200}{\hertz}. Uncertainties are one standard deviation.}
    \label{fig:sampling_rate}
\end{figure}

In Figure~\ref{fig:sampling_rate}, the experimental data is resampled to a wide range of sampling rates to assess the robustness of RING w.r.t. the sampling rate. RING achieves a nearly constant orientation estimation accuracy which only, unsurprisingly, degrades slightly for low sampling rates.

\begin{table}[hb]
\begin{center}
\caption{Ablation Study of the impact of individual IMT challenges on RING's experimental orientation estimation accuracy. In all scenarios, RING uses only magnetometer-free or 6D IMUs. Ablation study conducted at a sampling rate of \qty{100}{\hertz}.}
\label{tab:ablation}
\begin{threeparttable}
\begin{tabular}{cccc}
Nonrigid & S. Misal.\textsuperscript{1} & Sparse & MAE [deg] \\ \toprule
\xmark & \xmark & \xmark & $3.85\pm0.25$\\
\xmark & \xmark & \cmark & $3.69\pm0.22$\\
\xmark & \cmark & \xmark & $4.00\pm0.17$\\
\xmark & \cmark & \cmark & $4.36\pm0.49$\\ \midrule
\cmark & \xmark & \xmark & $6.13\pm0.57$\\
\cmark & \xmark & \cmark & $5.89\pm1.25$\\
\cmark & \cmark & \xmark & $7.38\pm0.50$\\
\cmark & \cmark & \cmark & $8.10\pm1.19$\\ \bottomrule
\end{tabular}
\begin{tablenotes}
\scriptsize \normalfont
\item \textsuperscript{1} Sensor-to-segment Misalignment
\end{tablenotes}
\end{threeparttable}
\end{center}
\end{table}

\subsection{Real-time Applicability of RING}
By design, RING can be applied online as it is defined by a step function (see eq.~\eqref{eq:ring}) that, based on the measurements at a certain timestep (see eq.~\eqref{eq:x_t}), returns an updated internal state and the rotational state estimate (see eq.~\ref{eq:y_t}) of the KC.
Therefore, if the step function executes faster than the sampling rate requires, then it is said to be real-time capable.
After compilation, the runtime of the step function of RING is $(794\pm16.7)\qty{}{\SIUnitSymbolMicro s}$ on a M2 Macbook Pro.
Thus, RING is real-time capable up to $\approx \qty{1000}{\hertz}$.

\section{Conclusion}
In this work, we have extended RING, a powerful IMT method, to generalize across a wide range of sampling rates, and we have showcased that it can simultaneously overcome the four key challenges in IMT: inhomogeneous magnetic fields, sensor-to-segment misalignment, sparse sensor setups, and nonrigid sensor attachment.
With an experimental tracking MAE of $8.10\pm 1.19$ degrees if all four challenges are present simultaneously, RING accurately estimates the rotational state of a three-segment KC from IMU measurements.
RING leverages a decentralized network of message-passing RNNs that is trained on simulated data but is capable of zero-shot generalization to real-world data.
Our evaluations reveal RING's superiority over SOTA methods in terms of accuracy and applicability, additionally, we demonstrate RING's robustness across various sampling rates, and its real-time capability.
By enabling plug-and-play usability and extending the applicability of inertial motion capture technology, RING not only advances the field but also opens new avenues for research and practical applications in environments previously deemed challenging.

\begin{ack}
The authors gratefully acknowledge the scientific support and HPC resources provided by the Erlangen National High Performance Computing Center (NHR@FAU) of the Friedrich-Alexander-Universität Erlangen-Nürnberg (FAU). The hardware is funded by the German Research Foundation (DFG).
\end{ack}

\bibliography{ifacconf}

\end{document}